\documentclass{article}
\usepackage{ijcai22}

\usepackage{times}
\usepackage{soul}
\usepackage{url}
\usepackage[hidelinks]{hyperref}
\usepackage[utf8]{inputenc}
\usepackage[small]{caption}
\usepackage{graphicx}
\usepackage{amsmath}
\usepackage{amsthm}
\usepackage{booktabs}
\usepackage{algorithm}
\usepackage{algorithmic}
\usepackage{amssymb}
\usepackage{amsthm}
\usepackage{multirow}
\usepackage{subfigure}
\usepackage{color}

\newtheorem{theorem}{Theorem}

\title{Stabilizing and Enhancing Link Prediction through Deepened Graph Auto-Encoders}

\author{Xinxing Wu, Qiang Cheng{\footnote{Corresponding author.}}
\affiliations
University of Kentucky, Lexington, Kentucky, U.S.A.
\emails
 xinxingwu@gmail.com,
 qiang.cheng@uky.edu
}

\begin{document}

\maketitle

\begin{abstract}
Graph neural networks have been widely used for a variety of learning tasks. Link prediction is a relatively under-studied graph learning task, with current state-of-the-art models based on one- or two-layer shallow graph auto-encoder (GAE) architectures. In this paper, we overcome the limitation of current methods for link prediction of non-Euclidean network data, which can only use shallow GAEs and variational GAEs. Our proposed methods innovatively incorporate standard auto-encoders (AEs) into the architectures of GAEs to capitalize on the intimate coupling of node and edge information in complex network data. Empirically, extensive experiments on various datasets demonstrate the competitive performance of our proposed approach. Theoretically, we prove that our deep extensions can inclusively express multiple polynomial filters with different orders. The codes of this paper are available at https://github.com/xinxingwu-uk/DGAE.
\end{abstract}

\section{Introduction}
Deep neural networks (DNNs) are effective in dealing with regular or Euclidean data, but they become ineffective for irregular or non-Euclidean data, such as citations of academic papers~\cite{Sen2008} and hyperlinks of web pages on the world wide web~\cite{Pei2020,Rozemberczki2020}. These data, typically irregular, complex, and of considerable dependencies, are often described in terms of graphs, posing notable difficulties to standard neural networks (NNs) designed for regular or Euclidean data. To model the interrelationship and structures of these data, graph neural networks (GNNs)~\cite{Scarselli2009} have attracted a thrust of research attention~\cite{Kipf2017} for extending standard NNs to learn over graph-structured data through iterative message passing processes between nodes and their neighbors.

\begin{figure*}[t]
\begin{center}
\subfigure[GAEs without features of nodes]{
\centering
{
\begin{minipage}[t]{0.4\linewidth}
\centering
\centerline{\includegraphics[width=1.1\textwidth]{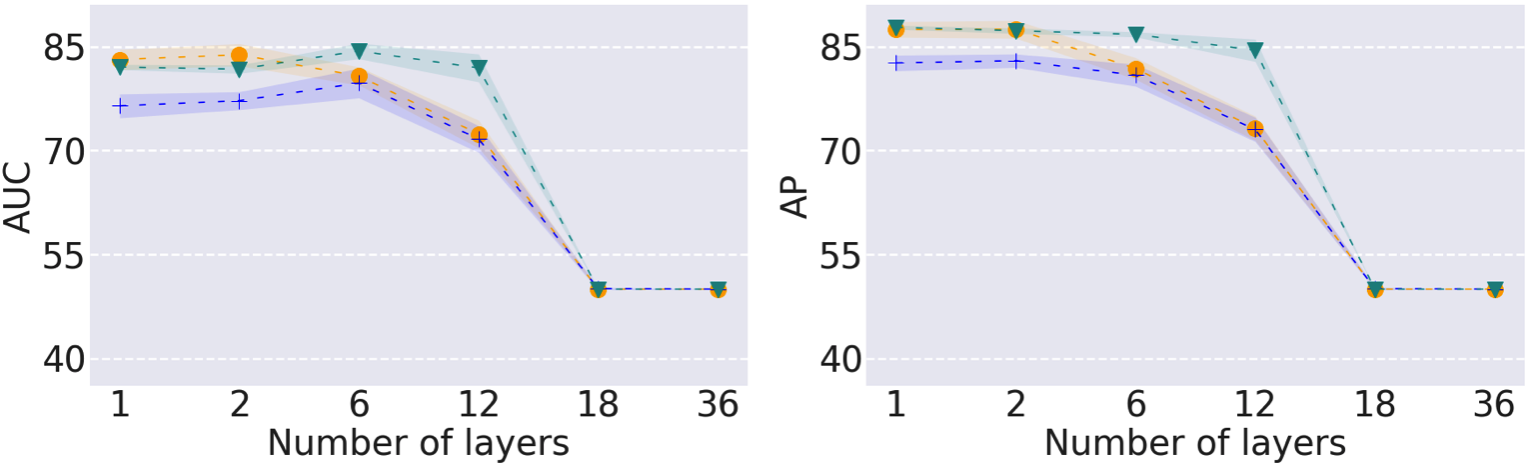}}
\end{minipage}%
}%
}%
\hspace{0.6in}
\subfigure[GAEs with features of nodes]{
\centering
{
\begin{minipage}[t]{0.4\linewidth}
\centering
\centerline{\includegraphics[width=1.1\textwidth]{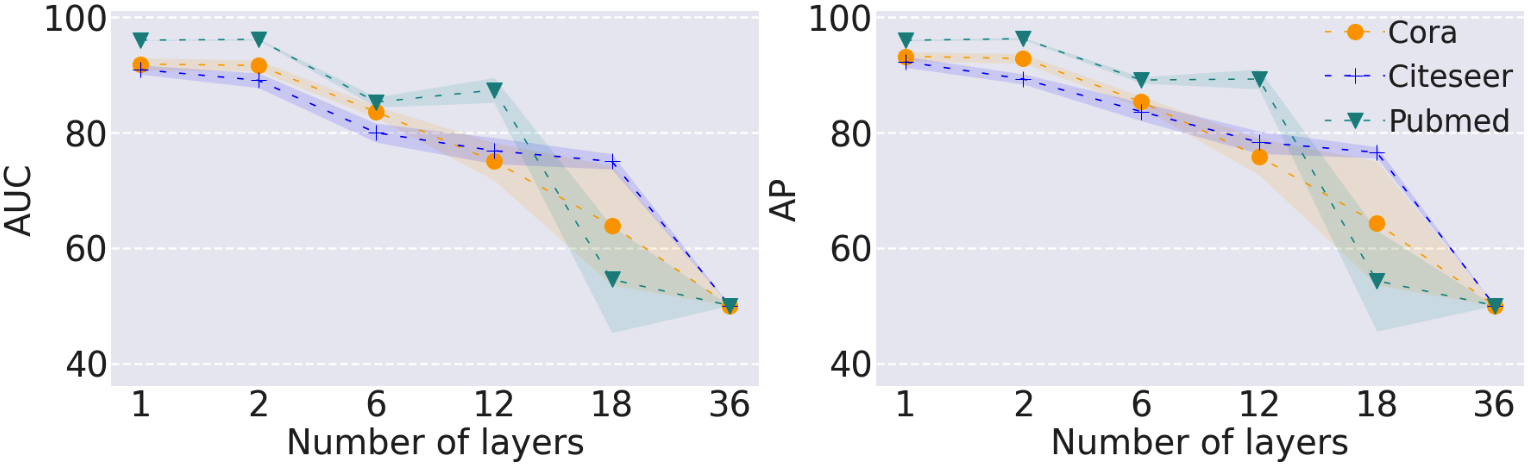}}
\end{minipage}%
}%

}%
\end{center}
\vskip -0.2in
\caption{Link prediction results in the area under the receiver operating characteristic (ROC) curve (AUC) and the average precision (AP) of standard GCN-based GAEs with different numbers of layers ($k=1, 2, 6, 12, 18$, and $36$). The results are averaged over $10$ runs of $10$ different random splits of datasets. It is observed that the performance of GAEs degrades with the increase of layers.}
\vskip -0.2in
\label{fig:01}
\end{figure*}

Link prediction is an important problem in graph-structured data analysis that predicts whether two nodes in a graph are likely to be linked~\cite{David2007}. It has a wide range of applications, for instance, reconstruction of metabolic networks, recommendation of hyperlinks of web pages, citations of academic papers, and friend recommendations in Twitter and Facebook. Graph convolutional networks (GCNs)-based graph auto-encoders (GAEs) or variational graph auto-encoders (VGAEs)~\cite{Wang2016,Kipf2016} are typical unsupervised node embedding approaches to link prediction on graph-structured data. Compared to popular DNNs such as convolutional neural networks (CNNs)~\cite{LeCun1998} and ResNet~\cite{He2016}, GAEs and their variants~\cite{Pan2018,Grover2019} come with quite shallow architectures, often one or two layers. Recently,~\cite{Salha2020} has pointed out that the reason for using shallow structures is because linear cases (i.e., one-layer versions) of GAEs have already achieved impressive performance and often outperform their multi-layer versions, including two- or three-layer GAEs-based models, on link prediction. Indeed, our experiments on real data also demonstrate that directly increasing the depth of GAEs does not help improve link prediction performance, but rather decreases it (Figure~\ref{fig:01}). Considering the potential of DNNs with many layers in various applications, the ability of these shallow GAEs-based models to extract information from higher-order neighborhoods and node features is patently limited.

Recently, a similar limitation has also been observed in the (semi-) supervised learning with GNNs. For example, different models for node classification,  such as GCN~\cite{Kipf2017}, GAT~\cite{Velickovic2018}, and EGNN~\cite{Gong2019}, usually achieved their best performance with two-layer structures. When stacking more layers, the performance of these models tends to degrade. Such a phenomenon has been referred to as over-smoothing where, as the number of layers increases, the node representation in these models tends to converge to a certain value and thus become indistinguishable~\cite{Kipf2017}. Lately, a number of studies have been conducted to address the over-smoothing problem of (semi-) supervised learning with GNNs, e.g.,~\cite{Xu2018,Li2018,Rong2020,Chen2020}.

In general, GNNs-based node classification, such as~\cite{Kipf2017}, is (semi-) supervised learning, and its final output is a multi-dimensional vector consisting of predicted labels for unlabeled nodes. In contrast, link prediction, such as~\cite{Kipf2016}, is typically regarded as an unsupervised learning task, and its final output is an adjacency matrix of a much higher dimension than that of the nodes. Thus, link prediction can be regarded as an inverse problem of decoding a noisy adjacency matrix with potentially many missing values of a graph into the underlying true values and thus appears to be more challenging than node classification in terms of supervision and potential output dimensionality. Besides, although the performance of both node classification and link prediction degrades with the increase of the number of layers, due to the nature of GNNs, the accuracies of node classification eventually converge to different constants~\cite{Wang2019,Chen2020}, while the final results of link prediction converge to $0.5$ (see Figure~\ref{fig:01}). Given all these patent differences, although the recent development in GNNs makes a deeper extension of GCN for node classification feasible, whether there is an advantage of increasing the depth of GAEs (VGAEs) for link prediction still remains unanswered.

In this paper, we study the deep extension of current GAE (VGAE) for link prediction, effectively extending it into a deep model, DGAE (DVGA).{\footnote{Due to page limitations, the properties and empirical experiments of DVGA are contained in our extended version at https://arxiv.org/abs/2103.11414}} To this end, the multi-scale information of seamlessly integrated adjacency matrix and node features will play a key role. Specifically, we create an approach to exploit the multi-scale information by constructing residual connections from the low-dimensional representations of the coherently fused adjacency matrix and node features of the graph. As the low-fidelity input graph consists of noisy or missing connections and the node features may also contain useless or irrelevant features, the input graph is intrinsically low-dimensional in both graph structure and node features. To capture such an essential, low-dimensional structure and provision it at different scales, we generate the co-embedding of the integrated adjacency matrix and node features by leveraging standard one-layer auto-encoders (AEs), and then use it to produce multi-scale information via skip connections, thereby further empowering the compact joint edge-node representation. To prevent the model from overfitting or going unnecessarily complex, we also adopt helpful regularizations. In brief, the contributions of the paper are summarized as follows:
\begin{itemize}
\item Our approach effectively extends the shallow GAE into the deep case, DGAE. To our best knowledge, it is the first deep GAE-based model for link prediction. 
\item Our approach innovatively incorporates standard AEs and graph AEs into unified models. Standard AEs are leveraged to obtain the low-dimensional representation of the seamlessly integrated adjacency matrix and node features, forming the basis for constructing conducive skip connections in graph AEs to extract multi-scale joint edge-node information.
\item Empirical experiments demonstrate that DGAE achieves superior performance to current shallow GAEs and their variants consistently across benchmarking datasets.\item Theoretical results show that DGAE can inclusively express multiple polynomial filters with different orders. 
\end{itemize}

The notations are used as follows. Let $G=(\mathbf{V},\mathbf{E}, \mathbf{X})$ be an undirected graph, where $\mathbf{V}$ and $\mathbf{E}$ respectively denote the sets of nodes and edges, and $\mathbf{X}\in\mathbb{R}^{n\times m}$ is the node feature matrix, with $n$ and $m$ being the numbers of nodes and features, respectively. $\mathbf{A}\in{\{0, 1\}}^{n\times n}$ is the adjacency matrix associated with $G$. $\|\cdot\|_{\mathrm{F}}^2$ denotes the Frobenius norm, and $\mathbb{E}_{\mathbf{A}}\left[\cdot\right]$ is the expectation over the distribution of $\mathbf{A}$.

\section{Related Work}
In this section, we briefly review the relevant studies in deepening models for GCNs; then, we discuss the developments of link prediction based on GAEs (VGAEs).

\paragraph{Deepening GCNs for Node Classification.} The deep extension of standard NNs remarkably improves their learning performance, leading to a quantum leap in artificial intelligence and applications. For GCNs, the over-smoothing phenomenon impedes the development of deep extensions of GCNs, whose performance degrades with the increasing number of layers. Currently, two major types of efforts have been taken to gain deeper GCNs: The first uses the residual connection~\cite{He2016}, a now-standard, proven-effective technique for training DNNs. For example,~\cite{Kipf2017} used residual connections between hidden layers;~\cite{Xu2018} introduced the first deep GCN, JKNet, which concatenated the node representations in all previous layers. Although highly effective for DNNs over regular data,~\cite{Xu2018,Kipf2017} showed that adding residual connections in GCNs only alleviates the performance degradation as the depth increases, with the resultant best performance still dominated by shallow GCN models. Recently,~\cite{Chen2020} proposed a model called GCNII, which combined a skipping connection from the input node features to each layer and adopted the identity mapping as a regularization. GCNII showed good performance on node classification with deepened GCNs. The second type uses deep propagation to capture higher-order information in graph-structured data. However, these models~\cite{Klicpera2019,Wu2019} are typically linear combinations of neighbor information in different layers, so in nature they are still regarded as shallow models~\cite{Chen2020}.

Besides the techniques focusing on architectures, two main regularizations are adopted in deepening GCNs: 1) Identity mapping~\cite{Hardt2017}. For example,~\cite{Chen2020} added an identity matrix to the weights of different layers of GCNII for regularization. 2) Dropout~\cite{Hinton2012}. As a standard regularization technique in DNNs, it was also used for mitigating the effects of over-smoothing in GCNs. For instance,~\cite{Rong2020} randomly dropped out edges to decelerate the performance degradation in deep GCNs for node classification. 

In this paper, noticing its positive effects on deepening DNNs and GCNs for node classification, we will leverage the residual connection to deepen GAEs-based models for link prediction with potentially useful regularization incorporated. It is worth noting that GAEs-based link prediction is patently different from GCNs-based node classification. The former, as an unsupervised learning technique, mainly reconstructs the underlying true graph based on a low-fidelity version of the graph structure and node features with potentially many missing edges, and its final output is a full $n\times n$ matrix. In contrast, GCNs are (semi-) supervised learning models typically regard the adjacency matrix as reliably given in predicting node labels, finally outputting a label for each unlabeled node. Thus, deepening GAEs-based models for link prediction appears to be more challenging than deepening GCNs for node classification, especially regarding how to effectively harness node features and noisy links.

\paragraph{GAEs (or VGAEs)-based Link Prediction.} GAEs- and VGAEs-based link prediction encodes the node features and the noisy topological structure with potentially missing edges in a graph into a latent representation and then predict the likelihood of an association between every pair of nodes.

Two-layer GAEs and VGAEs for link prediction were firstly proposed in~\cite{Kipf2016}, followed by a series of developments of graph representations. By incorporating an adversarial training module,~\cite{Pan2018} introduced two variants, adversarially regularized graph auto-encoders (ARGAs) and adversarially regularized variational graph auto-encoders (ARVGAs), both requiring the matching of the underlying representation with a specific prior distribution while minimizing the reconstruction error of the graph structure. \cite{Grover2019} improved the decoder in GAEs and VGAEs with an iterative graph refinement strategy, resulting in Graphite-AEs and Graphite-VAEs. In a recent study, ~\cite{Salha2020} pointed out that linear cases of GAEs and VGAEs already achieved a good performance compared to multi-layer cases, such as two- and three-layer GAEs- and VGAEs-based models, on link prediction. This study also questioned the relevance of using three citation network datasets, including Cora, Citeseer, and Pubmed, to compare complex GAEs and VGAEs. The architectures of all these models are generally not sufficiently deep; for example, in \cite{Pan2018}, a two-layer encoder was employed with a two-layer discriminator; in \cite{Grover2019}, a two-layer encoder and a three-layer decoder were combined with reverse message passing; in~\cite{Salha2020}, one-layer, two-layer, and three-layer GAEs and VGAEs were analyzed.

For simplicity, this paper will focus on extending the original GAEs; that is, our deepening is mainly for the encoder while retaining the original inner product architecture of the decoder. We will compare our deepened model with the shallow GAEs and their variations. Besides, we will also compare with Spectral Clustering (SC)~\cite{Tang2011} and DeepWalk~\cite{Perozzi2014} as two baselines.

\section{Proposed Model}
In this section, we will present our deep extension, DGAEs.

\paragraph{An Attempt to Directly Deepen GAEs.} GAE consists of an encoder and a decoder. It utilizes low-fidelity topological structure information encoded in $\mathbf{A}$ and $\mathbf{X}$ as a combined input to obtain a latent representation $\mathbf{Z}$; the decoder then reconstructs an estimated $\hat{\mathbf{A}}$ from $\mathbf{Z}$. Because the adjacency matrix of the graph represents the similarity or association between the pairs of nodes, the decoder for link prediction usually adopts an inner product of the learned latent representation, $\mathbf{Z}\mathbf{Z}^{\mathrm{T}}$. For simplicity, firstly we formalize the definition of a deepened $k$-layer GAE as follows:
\vskip -0.1in
\begin{equation}\label{generaldeepmodel}
\hat{\mathbf{A}}=\sigma(\mathbf{Z}\mathbf{Z}^{\mathrm{T}}),\,\,{\mathbf{Z}}=\tilde{\mathbf{A}}f_{k-1}(\cdots f_{1}(\mathbf{X}))\mathbf{W}_k,
\end{equation}
where $\hat{\mathbf{A}}$ is a reconstructed adjacency matrix as the output of link prediction, $\mathbf{Z}\in\mathbb{R}^{n\times h}$, $h$ is the dimension of the latent representation, $\sigma(\cdot)$ is an activation function, and $\tilde{\mathbf{A}}=\mathbf{D}^{-\frac{1}{2}}(\mathbf{A}+\mathbf{I})\mathbf{D}^{-\frac{1}{2}}$. Here, $\mathbf{D}$ is a diagonal matrix with $\mathbf{D}_{ii}=\sum_{j}(\mathbf{A}+\mathbf{I})_{ij}$, and $\mathbf{I}$ is the identity matrix. In this paper, we use the logistic sigmoid function for the output layer as~\cite{Kipf2016} and ReLU for other layers. Thus, we have $f_{i}(\mathbf{X})=\mathrm{ReLU}(\tilde{\mathbf{A}}\mathbf{X}{\mathbf{W}}_i)$, $i\geqslant 1$. For $i=0$, we take $f_{0}(\mathbf{X})=\tilde{\mathbf{A}}\mathbf{X}\mathbf{W}_1$. It is clear that we have the following special cases: \eqref{generaldeepmodel} respectively reduces to the linear and three-layer cases~\cite{Salha2020} with $k=1$ and $3$, and to GAE~\cite{Kipf2017} with $k=2$.

For model~\eqref{generaldeepmodel}, we train it by optimizing the following reconstruction error as a cost function between $\mathbf{A}$ and $\hat{\mathbf{A}}$, which was also used in~\cite{Kipf2016}:
\vskip -0.1in
\begin{equation}
\mathcal{L}_1\triangleq-\mathbb{E}_{\mathbf{A}}\left(\mathrm{log} \hat{\mathbf{A}}\right)=-\sum_{i,j}\mathbb{E}_{a_{ij}}\left(\mathrm{log} \hat{a}_{ij}\right),
\end{equation}
where $a_{ij}$ is the ($i$,$j$)-element of $\mathbf{A}$ for training. $\hat{a}_{ij}=\sigma(\mathbf{z}_i\mathbf{z}_j^{\mathrm{T}})$ predicts the probability of a link between nodes $i$ and $j$ with $\mathbf{z}_i$ ($\mathbf{z}_j$) the $i$-th ($j$-th) row-vector of $\mathbf{Z}$.

For $k$-layer GAE, we compute the cases of $k=1$ $(\mathrm{linear\,\,case}), 2$ $(\mathrm{GAE}), 6, 12, 18$, and $36$ over three standard datasets, with the results shown in Figure~\ref{fig:01}. It is seen that the performance of link prediction, in AUC and AP, decreases with the increase of depth. Interestingly, it is observed that on all three datasets, when the number of layers reaches $36$, all results degenerate to $0.5$. With $\tilde{\mathbf{A}}=\mathbf{D}^{-\frac{1}{2}}(\mathbf{A}+\mathbf{I})\mathbf{D}^{-\frac{1}{2}}$, the GAE model corresponds to a lazy random walk, where each time the walker stays at the current node with a probability of
0.5 or walks to a random neighbor. Thus, the link prediction with GAE will finally converge to $0.5$ with the increase of layers.

\paragraph{Skip Connection of Node Features.} In Figure~\ref{fig:01}, it is noted that good performance is generally achieved by models with one or two layers, implying that the information from the shallow layers will play a crucial role in the final performance. In the development of deepened GCNs for node classification, residual connections are used to retain the information in previous layers; for example, ~\cite{Chen2020} added the input node features or the output of a fully connected neural network on node features to each layer. For link prediction, the (potentially high-dimensional) node features may contain irrelevant features and are usually noisy. Inspired by the observation drawn from Figure~\ref{fig:01}, we propose to capture the essential, low-dimensional subspace of the input node features. Specifically, we propose to incorporate standard AEs on node features to obtain latent low-dimensional representations at different layers and fuse them using skip connections in~\eqref{generaldeepmodel}, thereby maximally preserving the useful node information in previous layers:
\vskip -0.15in
\begin{equation}\label{residualdeepmodel1}
\left\{\begin{array}{l}
\hat{\mathbf{A}}=\sigma(\mathbf{Z}\mathbf{Z}^{\mathrm{T}}),\\
{\mathbf{Z}}=\alpha_{k-1}\tilde{\mathbf{A}}f_{k-1}(\cdots(\alpha_1 f_{1}(\mathbf{X})+(1-\alpha_1)g_1(\mathbf{X})))\mathbf{W}_k\\
\,\,\,\,\,\,\,\,\,\,\,\,+(1-\alpha_{k-1})\tilde{\mathbf{A}}g_{k-1}(\mathbf{X})\mathbf{W}_k,\\
g_i^{\prime}(g_i(\mathbf{X}))=\mathbf{X}, i=1,\ldots,k-1,
\end{array}\right.
\end{equation}
where $g_i$ and $g_i^{\prime}$ are respectively the encoder and decoder of standard AEs, with the dimension of $g_i^{\prime}$ being the same as $f_{i}$, and $0\leqslant\alpha_i\leqslant1\,(i=1,\ldots,k-1)$ are combination coefficients.

For optimizing~\eqref{residualdeepmodel1}, we minimize the following combined reconstruction errors:
\vskip -0.15in
\begin{equation}\label{residualdeepmodel1loss}
\mathcal{L}_2=\sum_{i=1}^{k-1}\lambda_i\|g_i^{\prime}(g_i(\mathbf{X}))-\mathbf{X}\|_{\mathrm{F}}^2+\lambda_0\mathcal{L}_1,
\end{equation}
where $\lambda_i\geqslant0 \,(i=0,\ldots,k-1)$ are hyper-parameters balancing the reconstruction errors of AEs with graph AEs.

\begin{figure}
\begin{center}
\centerline{\includegraphics[width=0.42\textwidth]{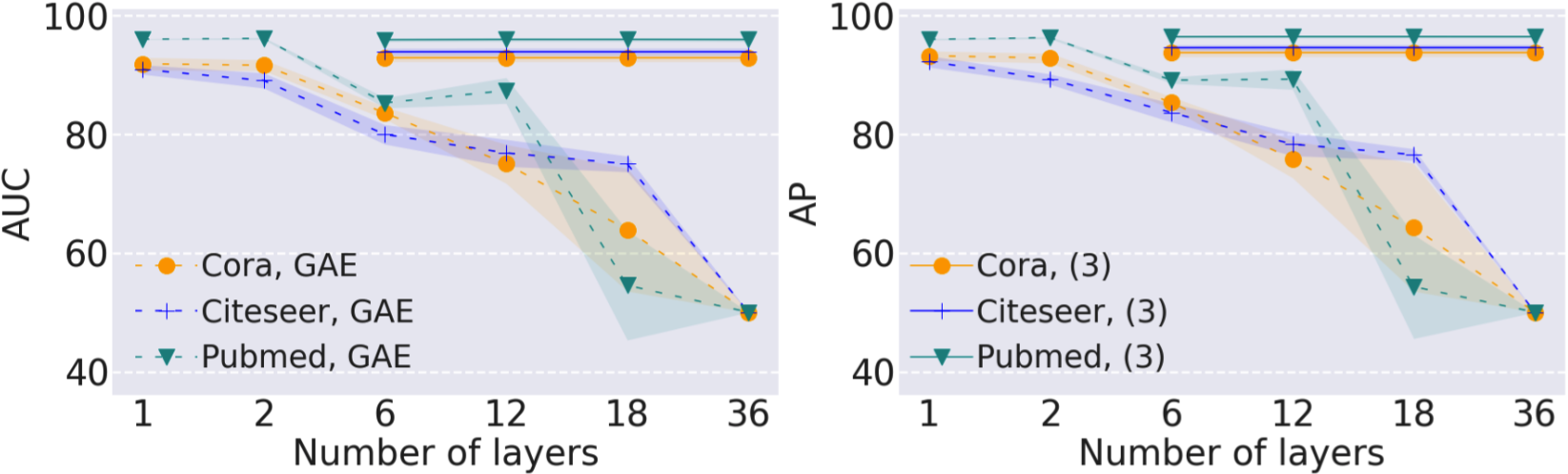}}
\end{center}
\vskip -0.25in
\caption{Comparison of model~\eqref{residualdeepmodel1} with $k$-layer GAEs~\eqref{generaldeepmodel}. The results are averaged over $10$ runs of $10$ different random splits of datasets. For $k=1, 2$, the results for \eqref{residualdeepmodel1} and $k$-layer GAEs are essentially the same (and thus not depicted for~\eqref{residualdeepmodel1} for visual clarity). Note that the legends are for both sub-figures.}
\vskip -0.2in
\label{fig:02}
\end{figure}

For model~\eqref{residualdeepmodel1}, we compute the cases of $k=6, 12, 18$, and $36$, and compare the results with those obtained using $k$-layer GAEs in Figure~\ref{fig:02}. It is found that model~\eqref{residualdeepmodel1} can effectively avoid the performance degradation when increasing the number of layers in model~\eqref{generaldeepmodel}; in particular, the improvement over Citeseer is substantial. Further, the standard deviations for model~\eqref{residualdeepmodel1} are much smaller than their GAE counterparts.

\paragraph{Integrating Adjacency Information and Node Features.} Model~\eqref{residualdeepmodel1} can exploit the low-rank subspace of potentially high-dimensional yet noisy node features, which may be sufficient for deepening the GCNs for node classification. However, the final output of link prediction is directly concerned about adjacency information; thus, predicting links in GAEs is conspicuously different from the node classification of GCNs. In particular, the intrinsic structure of the low-fidelity input adjacency matrix is under-utilized in~\eqref{residualdeepmodel1}. This input matrix is usually noisy with potentially many missing values; yet, it often resides on an essential, low-rank subspace due to the clustering structure of connected nodes~\cite{Fortunato2010}. To exploit such a low-rank subspace, we propose to co-embed the node features and the input adjacency matrix. Specifically,  we enhance~\eqref{residualdeepmodel1} using the fused adjacency information and node features as input to standard AEs as follows:
\vskip -0.15in
\begin{equation}\label{residualdeepmodel2}
\left\{\begin{array}{l}
\hat{\mathbf{A}}=\sigma(\mathbf{Z}\mathbf{Z}^{\mathrm{T}}),\\
{\mathbf{Z}}=\alpha_{k-1}\tilde{\mathbf{A}}f_{k-1}(\cdots(\alpha_1 f_{1}(\mathbf{X})+(1-\alpha_1)g_1(\tilde{\mathbf{A}}\mathbf{X})))\\
\,\,\,\,\,\,\,\,\,\,\,\,\mathbf{W}_k+(1-\alpha_{k-1})\tilde{\mathbf{A}}g_{k-1}(\tilde{\mathbf{A}}\mathbf{X})\mathbf{W}_k,\\
g_i^{\prime}(g_i(\tilde{\mathbf{A}}\mathbf{X}))=\tilde{\mathbf{A}}\mathbf{X}, i=1,\ldots,k-1,
\end{array}\right.
\end{equation}
where the compact co-embedding is formed using standard AEs for nonlinearly capturing the low-rank subspace from the seamlessly integrated (convolved) adjacency matrix and node matrix. The co-embedding is then provided to various layers of our model via the dense skip connections. Correspondingly, we use the following cost function for optimization, which incorporates the reconstruction errors of using standard AEs for co-embedding: $\mathcal{L}_3=\sum_{i=1}^{k-1}\lambda_i\|g_i^{\prime}(g_i(\mathbf{\tilde{A}X}))-\mathbf{\tilde{A}X}\|_{\mathrm{F}}^2+\lambda_0\mathcal{L}_1$.

\begin{figure}
\begin{center}
\centering
\centerline{\includegraphics[width=0.42\textwidth]{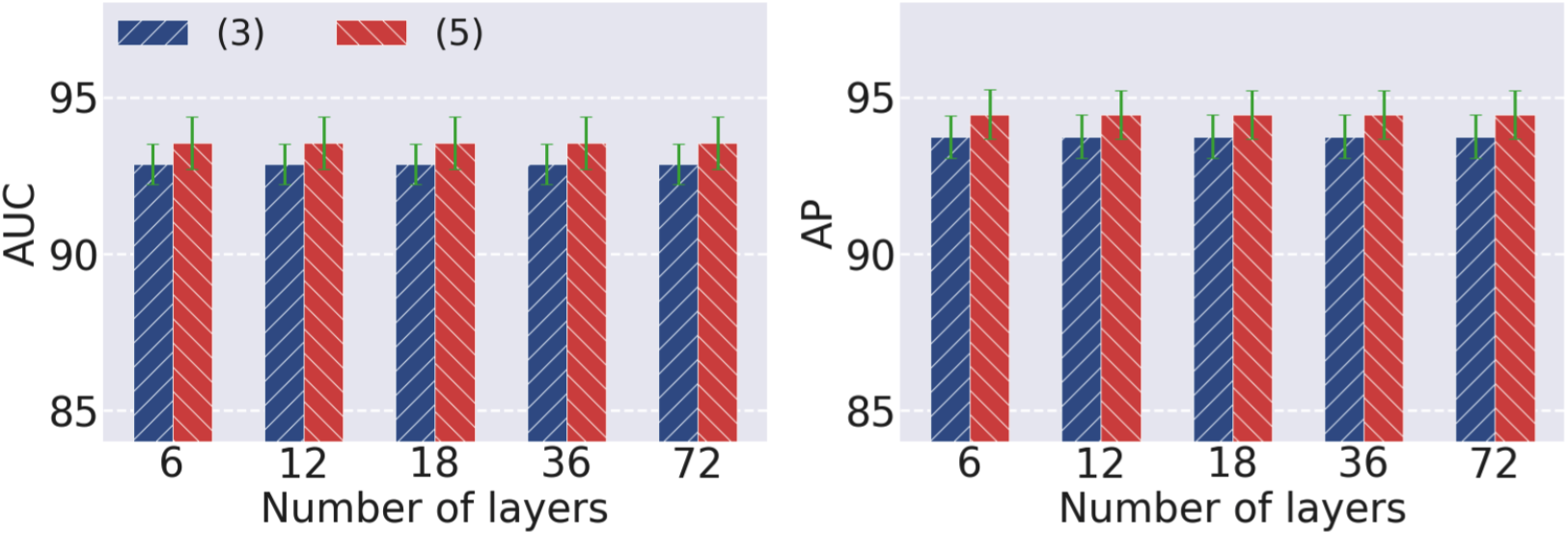}}
\end{center}
\vskip -0.3in
\caption{The comparison of models on Cora. The results are averaged over $10$ runs of $10$ different random splits.}
\label{Compare3and5}
\vskip -0.15in
\end{figure}

We empirically compare the models in~\eqref{residualdeepmodel1} and~\eqref{residualdeepmodel2} for $k=6, 12, 18$, $36$, and $72$; see Figure~\ref{Compare3and5}. Model~\eqref{residualdeepmodel2} evidently has a better performance than model~\eqref{residualdeepmodel1}; it demonstrates that capturing the low-rank subspace of the seamlessly integrated graph structure with node features and using it as input to various layers via the residual connections in GAEs is well conducive to the final link prediction.

\paragraph{Deep Model with Regularization.} Noting that the identity matrix~\cite{Hardt2017} was helpful in GCNII~\cite{Chen2020} for node classification as a regularization, we consider to add it as a regularization term in~\eqref{residualdeepmodel2} to potentially boost the performance. The change is mainly for the ${\mathbf{Z}}$ quantity in~\eqref{residualdeepmodel2}, and the modified version is as follows:
\vskip -0.15in
\begin{equation}\label{residualdeepmodel3}
\begin{array}{l}
{\mathbf{Z}}=\alpha_{k-1}\tilde{\mathbf{A}}f^{\prime}_{k-1}(\cdots (\alpha_1 f^{\prime}_{1}(\mathbf{X})+(1-\alpha_1)g_1(\tilde{\mathbf{A}}\mathbf{X})))\\
\,\,\,\,\,\,\,\,\,\,\,\,((1-\beta_k)\mathbf{W}_k+\beta_k\mathbf{I}_k)+(1-\alpha_{k-1})\tilde{\mathbf{A}}g_{k-1}(\tilde{\mathbf{A}}\mathbf{X})\\
\,\,\,\,\,\,\,\,\,\,\,\,((1-\beta_k)\mathbf{W}_k+\beta_k\mathbf{I}_k),
\end{array}
\end{equation}
where $f^{\prime}_{i}(\mathbf{X})=\mathrm{ReLU}(\tilde{\mathbf{A}}\mathbf{X}((1-\beta_i){\mathbf{W}}_i+\beta_i\mathbf{I}_i))$, and $\mathbf{I}_i$ is the identity matrix, $i=1,\ldots,k-1$. Besides, $0\leqslant\alpha_i\leqslant1\,(i=1,\ldots,k-1)$ and $0\leqslant\beta_i\leqslant1\,(i=1,\ldots,k)$ are combination coefficients. We denote~\eqref{residualdeepmodel2} by DGAE$_{\alpha}$ and~\eqref{residualdeepmodel3} by DGAE$_{\alpha}^{\beta}$.

\section{Experiments}{\label{Experiments}}
We perform experiments to verify our deep extension extensively by comparing it with shallow models and contemporary GAEs-based link prediction models on many benchmarking datasets.

\paragraph{Datasets Used.} 
Firstly, we employ three standard benchmark datasets, i.e., Cora, Citeseer, and Pubmed; then, we also evaluate our deep extensions on three webpage-related datasets. We summarize the data statistics in Table~\ref{datasets}.
\begin{table}
\begin{sc}
\begin{center}
\begin{tabular}{l|rrr}
\hline\hline
    Dataset&\# Nodes&\# Edges &\# Features\\
    \hline
    Cora		& 2,708 	& 5,429			& 1,433\\ 
    Citeseer	& 3,327 	& 4,732			& 3,703\\
    Pubmed		& 19,717 	& 44,338		& 500\\ 
    Chameleon   & 2,277 	& 36,101 		& 2,325\\ 
    Texas		& 183 		& 309 			& 1,703\\ 
    Wisconsin   & 251 		& 499 			& 1,703\\ 
\hline\hline
\end{tabular}
\end{center}
\end{sc}
\vskip -0.15in
\caption{Statistics of datasets.}
\label{datasets}
\vskip -0.2in
\end{table}

\paragraph{Experiment Settings.} In all experiments, we set the maximum number of epochs to $200$ and adopt the Adam optimizer with an initial learning rate of $0.01$. We follow the training scheme in~\cite{Kipf2016,Pan2018}: We train all the models by randomly removing $15\%$ of links while keeping all node features, and the validation and test sets are formed by a ratio of 5:10 from the removed edges and the corresponding node pairs; the models' weights are initialized by using the Glorot uniform technique. For simplicity, we construct our deep encoders with $(k-1)$ 32-neuron hidden layers for $k$ in~\eqref{residualdeepmodel2} or~\eqref{residualdeepmodel3} and a 16-neuron latent embedding layer. Besides, we perform a grid search in \{0.000001, 0.000005, 0.00001, 0.00005, 0.0001, 0.0005, 0.001, 0.005, 0.01, 0.05, 0.1, 0.5, 1, 1.5, 2\} to tune hyper-parameters for our models according to the performance on the validation set.

\paragraph{Evaluation Metrics.} We evaluate performance based on two metrics, AUC and AP, which were also adopted for link prediction in~\cite{Kipf2016,Grover2019,Pan2018}. We directly use their evaluation codes for fair comparison. For all experiments, the obtained mean results and standard deviations are for $10$ runs over $10$ different random train/validation/ test splits of datasets.

\begin{figure}[!thbp]
\begin{center}
\centering
\centerline{\includegraphics[width=0.42\textwidth]{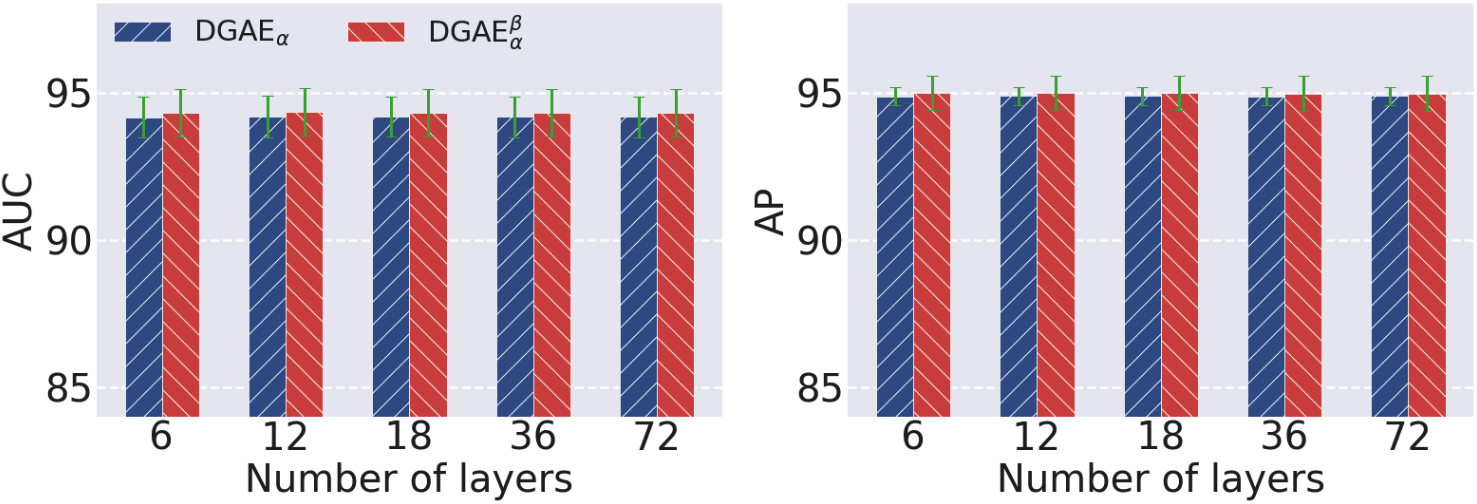}}
\end{center}
\vskip -0.3in
\caption{The comparison of DGAE$_{\alpha}^{\beta}$ with DGAE$_{\alpha}$ on Citeseer. The results are averaged over $10$ runs of $10$ different random splits.}
\label{FigureCompare5and6}
\vskip -0.15in
\end{figure}

\paragraph{Results on Cora, Citeseer, and Pubmed.} We summarize the AUC and AP results (mean$\pm$standard deviation) of our experiments in Table~\ref{results1}. Both $6$-layer and $36$-layer DGAE$_{\alpha}$ architectures achieve competitive results; in particular, our DGAE$_{\alpha}$ achieves consistently better predictive performance than the methods in comparison. Further, we compare DGAE$_{\alpha}^{\beta}$ with DGAE$_{\alpha}$ in Figure~\ref{FigureCompare5and6}. It is observed that DGAE$_{\alpha}^{\beta}$ achieves relatively better performance. 

\begin{table*}[!htbp]
\begin{small}
\begin{center}
\begin{tabular}{l|rrrrrr}
\hline\hline
\multirow{2}*{ Dataset}                     &\multicolumn{2}{c}{Cora}        &\multicolumn{2}{c}{Citeseer}        &\multicolumn{2}{c}{Pubmed}\\
    		                                & AUC (\%)		  &AP (\%)          & 	AUC (\%)	   &AP (\%)          & AUC (\%)		   &AP (\%) \\
\hline
    SC  		                            &$85.36\pm1.62$   &$88.41\pm1.18$   &$77.26\pm1.87$    &$80.24\pm1.36$   &$80.16\pm0.54$   &$79.89\pm0.66$   \\
    DeepWalk  		                        &$76.41\pm1.70$   &$81.78\pm1.02$   &$64.15\pm1.81$    &$74.60\pm1.21$   &$73.32\pm0.60$   &$81.53\pm0.40$    \\
    ARGA                                    &$92.10\pm1.18$   &$93.25\pm0.93$   &$90.43\pm0.73$    &$92.04\pm0.83$   &$95.55\pm0.21$   &$95.76\pm0.19$   \\
    Graphite-AE    		                    &$91.29\pm0.83$   &$92.63\pm0.70$   &$88.84\pm1.62$    &$89.64\pm1.60$   &$96.24\pm0.23$   &$96.42\pm0.21$    \\
    Linear-GAE  				                &$91.86\pm1.06$   &$93.20\pm0.80$   &$90.89\pm0.79$    &$92.24\pm0.95$   &$96.01\pm0.06$   &$95.98\pm0.11$   \\
	2-GAE 		                    &$91.61\pm0.92$	  &$92.81\pm0.82$   &$89.04\pm1.27$    &$89.26\pm0.87$   &$96.13\pm0.14$   &$96.26\pm0.17$   \\
	6-GAE 		                    &$83.58\pm1.01$	  &$85.32\pm0.97$   &$79.95\pm1.64$    &$83.63\pm1.59$   &$85.27\pm0.86$   &$89.10\pm0.54$   \\
	36-GAE 		                    &$50.00\pm0.00$	  &$50.00\pm0.00$   &$50.00\pm0.00$    &$50.00\pm0.00$   &$50.00\pm0.00$   &$50.00\pm0.00$   \\
	72-GAE 		                    &$50.00\pm0.00$	  &$50.00\pm0.00$   &$50.00\pm0.00$    &$50.00\pm0.00$   &$50.00\pm0.00$   &$50.00\pm0.00$   \\
\hline	
6-DGAE$_{\alpha}$		&{\bf\textit{93.55$\pm$0.84}}	  &{\bf\textit{94.46$\pm$0.78}}	&$94.16\pm0.69$	   &$94.86\pm0.30$	 &$96.51\pm0.16$   &{\bf\textit{96.79$\pm$0.15}}   \\
36-DGAE$_{\alpha}$			&$ 93.54\pm0.84$   &$94.45\pm0.78$   &{\bf\textit{94.17$\pm$0.68}}    &$94.87\pm0.30$   &{\bf\textit{96.52$\pm$0.15}}   &$96.79\pm0.16$     \\
72-DGAE$_{\alpha}$			&$93.53\pm0.85$			&$94.45\pm0.78$   & $ 94.17\pm0.69$     &{\bf\textit{94.88$\pm$0.29}}  &$ 96.51\pm0.16$   &$96.79\pm0.17$\\
\hline
(t-statistics, p-value)		&(3.17, 0.0054)					& (3.15, 0.0055)	  & 	(9.95, $<$0.0001) &(8.40, $<$0.0001)  & (3.22, 0.0047)  &	(4.53, 0.0003)		\\
\hline\hline
\end{tabular}
\end{center}
\end{small}
\vskip -0.15in
\caption{AUC (\%) and AP (\%) scores of different algorithms for link prediction. Here, the p-value represents the statistical significance of the difference between the best performance of our method (reported in the last 3 rows) and the best performance of other methods.}
\label{results1}
\vskip -0.05in
\end{table*}

\paragraph{Validation on Datasets of Webpage Networks.} While \cite{Salha2020} questioned the possibility of further improvement of using the three standard network datasets, Cora, Citeseer, and Pubmed, our deep extensions demonstrate clear competitive performance over current state-of-the-art (SOTA) algorithms on them; see Table~\ref{results1}. To further verify the performance of our deep extensions, we evaluate DGAE$_{\alpha}^{\beta}$ on three more datasets. The results in Table~\ref{results2} show that our deep extensions achieve superior performance to almost all algorithms in comparison. The gain is particularly pronounced on Chameleon because the edge information in Chameleon is richer than other datasets, implying that more edge information may yield more benefit through deepened learning.
\begin{table*}[!htbp]
\vskip -0.05in
\begin{small}
\begin{center}
  \begin{tabular}{l|rrrrrr}
    \hline\hline
\multirow{2}*{ Dataset} &\multicolumn{2}{c}{chameleon}                &\multicolumn{2}{c}{texas}       &\multicolumn{2}{c}{wisconsin} \\
    		            & AUC (\%)		&AP (\%)       &AUC (\%)	   &AP (\%)        & AUC (\%)		&AP (\%) \\
    \hline
Linear-GAE				   &$45.21\pm3.75$ &$53.74\pm3.79$ &$12.97\pm11.78$&$40.41\pm 8.45$&$21.98\pm 7.24$&$44.53\pm 5.64$ \\
2-GAE	                   &$46.33\pm6.26$ &$54.62\pm5.77$ &$22.19\pm12.65$&$46.35\pm 8.55$&$34.63\pm11.13$&$57.09\pm 7.12$ \\
6-GAE	                   &$37.67\pm5.80$ &$49.71\pm4.69$ &$15.31\pm10.66$&$42.38\pm 8.73$&$19.51\pm 7.88$&$45.14\pm 8.00$ \\
36-GAE	                   &$50.00\pm0.00$	  &$50.00\pm0.00$   &{\bf\textit{50.00$\pm$0.00}}    &$50.00\pm0.00$   &$50.00\pm0.00$   &$50.00\pm0.00$   \\
72-GAE	                   &$50.00\pm0.00$	  &$50.00\pm0.00$   &{\bf\textit{50.00$\pm$0.00}}    &$50.00\pm0.00$   &$50.00\pm0.00$   &$50.00\pm0.00$   \\
\hline
6-DGAE$_{\alpha}^{\beta}$  &$56.89\pm4.14$ &$62.89\pm4.42$ &$45.78\pm 9.63$&{\bf\textit{61.06$\pm$8.74}}&$53.95\pm10.92$&$67.76\pm 8.62$\\
36-DGAE$_{\alpha}^{\beta}$ &$58.35\pm6.61$ &$60.60\pm5.78$ &$45.16\pm10.86$&$60.44\pm 9.74$&{\bf\textit{54.20$\pm$10.55}}&{\bf\textit{67.78$\pm$8.73}}\\
72-DGAE$_{\alpha}^{\beta}$ &{\bf\textit{62.26$\pm$0.17}} &{\bf\textit{65.02$\pm$1.46}} &$ 46.09\pm 9.56$&$61.04\pm 8.82$&$53.70\pm 9.52$&$67.46\pm 7.80$\\
\hline
(t-statistics, p-value)		& 	(228.06, $<$0.0001)	& 	(5.52, $<$0.0001)	 & (-1.29,  0.2122)	&  	(4.00,	0.0008)	 &  (1.26, 0.2241) 	 &	 (3.00, 0.0077)		\\
    \hline\hline	
  \end{tabular}
\end{center}
\end{small}
\vskip -0.15in
\caption{AUC (\%) and AP (\%) scores for $k$-GAEs with $k=1, 2, 6, 36, 72$, and DGAE$_{\alpha}^{\beta}$ with layers $6$, $36$, and $72$. Here, the p-value represents the statistical significance of the difference between the best performance of our method (reported in the last 3 rows) and the best performance of other methods.}
\label{results2}
\vskip -0.2in
\end{table*}

\section{Discussions}
In this section, we provide discussions and further analysis of our proposed deep extensions.

\paragraph{Fusion of Standard and Graph AEs.} The combining of standard AEs in~\eqref{residualdeepmodel1} and~\eqref{residualdeepmodel2} may seem similar to the layer-wise pre-training technique in~\cite{Bengio2007,Hinton2006}, given that we train a series of shallow standard AEs to obtain multiple low-dimensional representations in an unsupervised fashion. But essential difference exists: The shallow standard AEs in our model are independent, and the extracted latent representations are used for every new skip connection in the entire deep GAEs, which are fused integrally. In contrast, for the layer-wise pre-training, multiple standard AEs are stacked and each low-dimensional representation is input to the next layer only. Moreover, from the results in Figure~\ref{Compare3and5}, it is evident that using standard AEs to jointly exploit the essential, low-rank subspace of the graph structure and node features for co-embedding is more conducive to the final link prediction than not using them. 

In addition, it is noted that our deep extension is different from~\cite{Wang2017}, which is based on pure AEs and only uses NNs up to 6 layers. In contrast, we establish a general ability to boost link prediction using deep GAEs with many layers (e.g., up to 72 layers in our experiments).

\begin{figure}[!tbhp]
\vskip -0.1in
\begin{center}
\subfigure[2-layer GAE]{
\centering
{
\begin{minipage}[t]{0.3\linewidth}
\centering
\centerline{\includegraphics[width=1.1\textwidth]{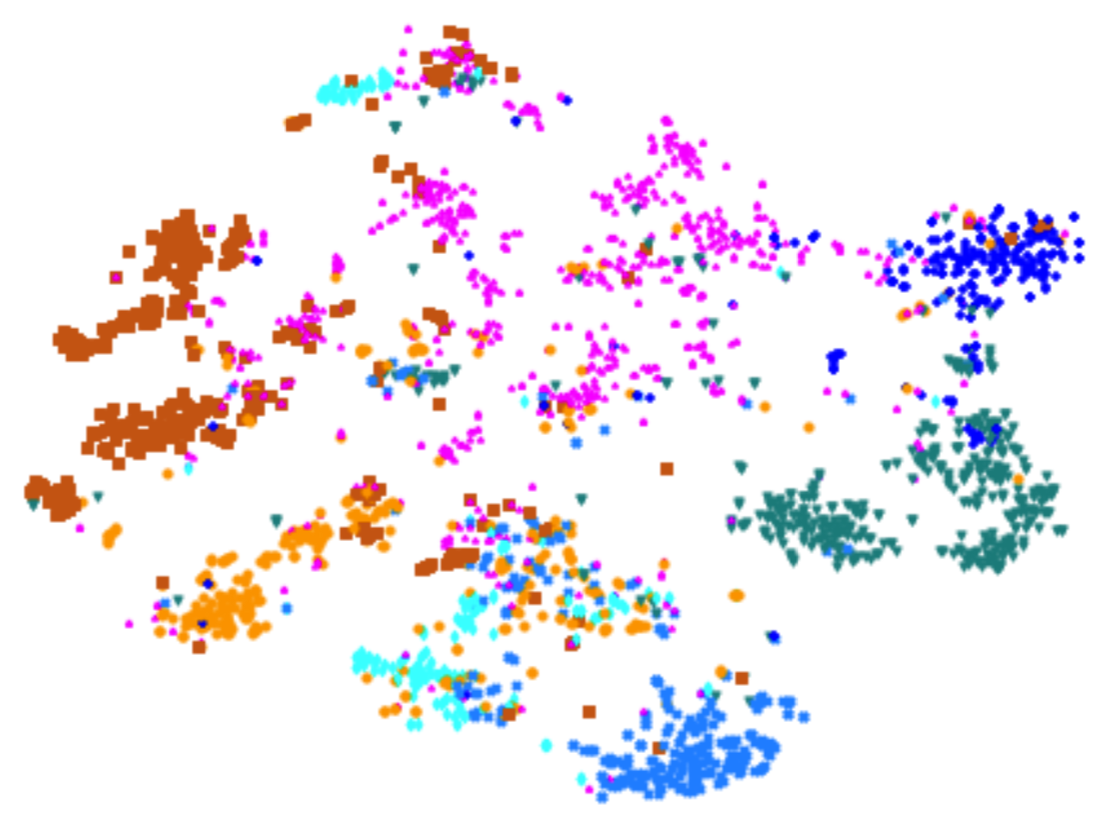}}
\end{minipage}%
}%
}%
\hspace{-0.01in}
\subfigure[36-layer GAE]{
\centering
{
\begin{minipage}[t]{0.3\linewidth}
\centering
\centerline{\includegraphics[width=1.1\textwidth]{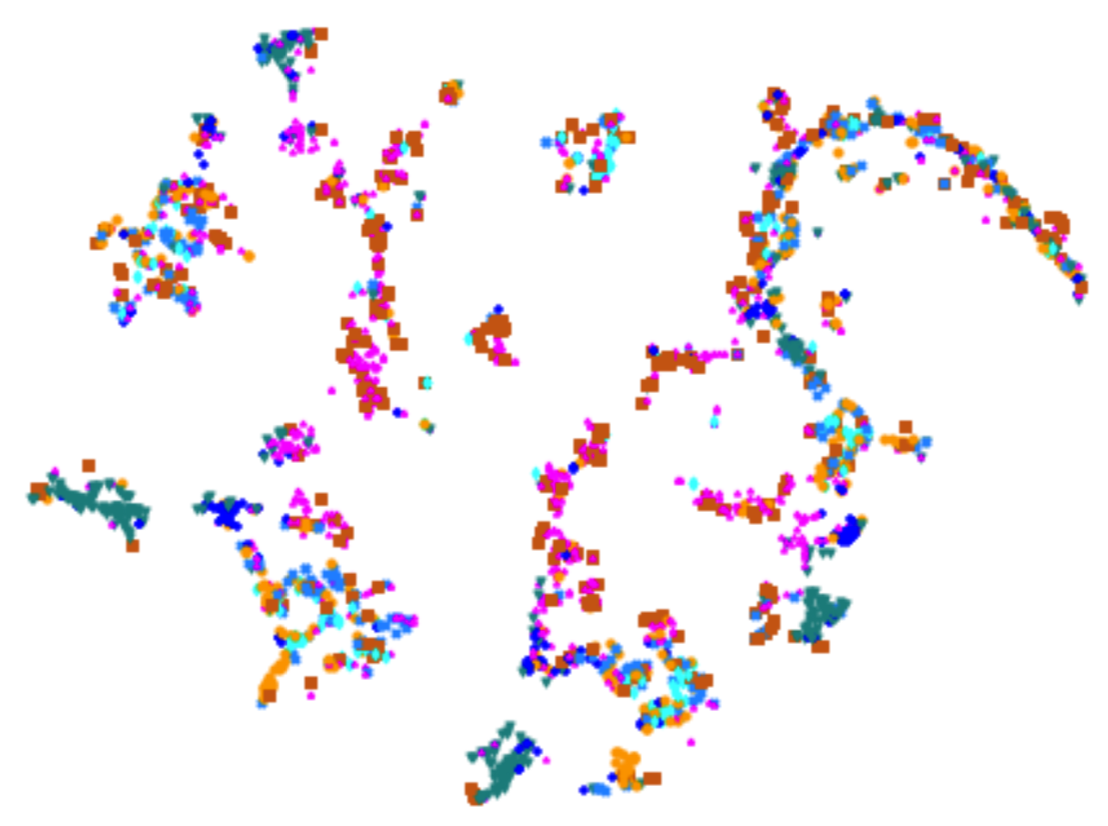}}
\end{minipage}%
}%
}%
\hspace{-0.01in}
\subfigure[36-layer DGAE$_{\alpha}^{\beta}$]{
\centering
{
\begin{minipage}[t]{0.3\linewidth}
\centering
\centerline{\includegraphics[width=1.1\textwidth]{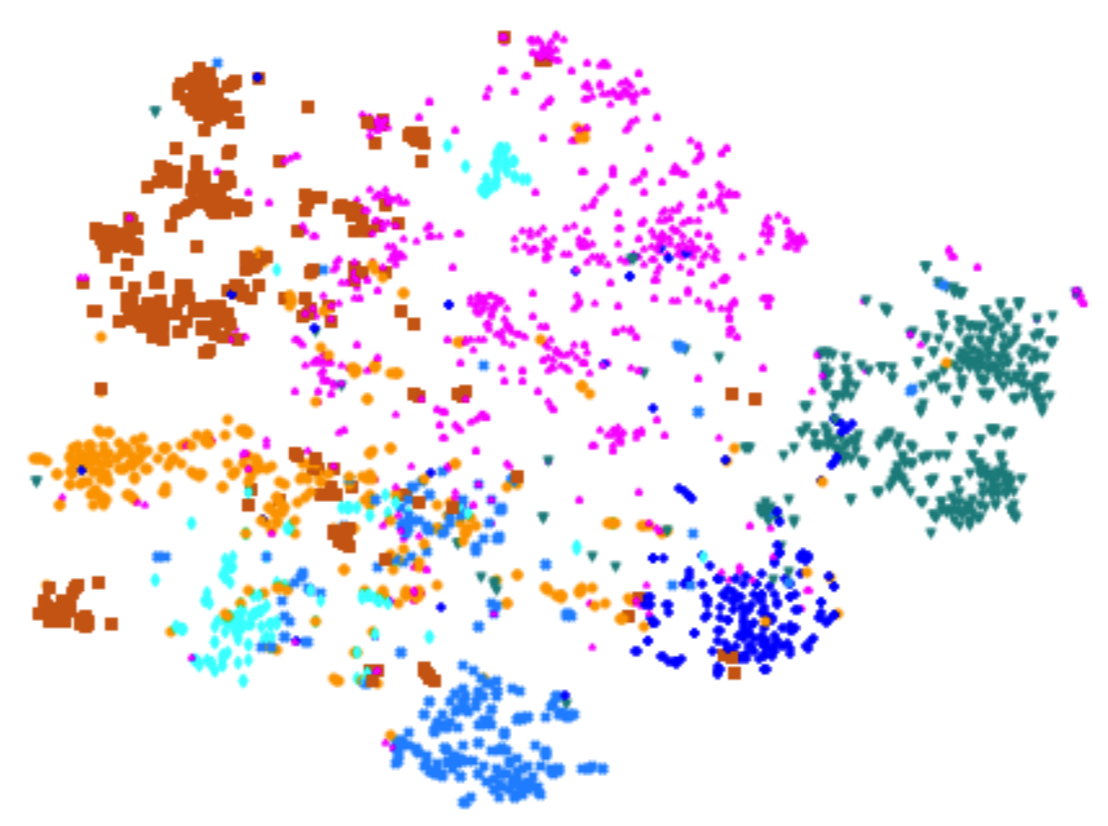}}
\end{minipage}%
}%
}%
\end{center}
\vskip -0.2in
\caption{Visualization of low-dimensional embeddings for Cora. Each color denotes a document class. During training, neither DGAE$_{\alpha}^{\beta}$ nor GAE has access to class information. (a) Silhouette: 0.125 and NMI: 0.432; (b) Silhouette: -0.045 and NMI: 0.033; (c) Silhouette: 0.169 and NMI: 0.482.}
\vskip -0.16in
\label{fig:results4}
\end{figure}

\paragraph{Visualization.} To understand the difference between the learning in DGAE$_{\alpha}^{\beta}$ and GAEs, we visualize the low-dimensional embeddings learned by DGAE$_{\alpha}^{\beta}$ and GAEs with different numbers of layers using 2D t-SNE~\cite{Maaten2008}. The 2D embedding results on Cora are visually presented in Figure~\ref{fig:results4}. It is observed that DGAE$_{\alpha}^{\beta}$ leads to more separated graph embeddings than GAEs, which are in line with our quantitative results in AUC and AP as shown above and the grouping quality in Silhouette coefficient and normalized mutual information (NMI).

\paragraph{Multiple Graph AEs with Different Depths.} Next, we analyze why our deep extension can stabilize and enhance link prediction. We theoretically show that our deep extension inclusively express multiple polynomial filters with different orders.

\begin{theorem}{\label{Thm1}}
A $k$-layer ($k\geqslant 3$) DGAE$_{\alpha}$ is a linear combination of $2$-order to $k$-order polynomial filters with arbitrary combination coefficients.
\end{theorem}
\begin{proof}
Note that, for ${\mathbf{Z}}$ in~\eqref{residualdeepmodel2}, $\forall k\geqslant3$, we have
$$
\begin{array}{rl}
{\mathbf{Z}}\,=&\alpha_{k-1}\tilde{\mathbf{A}}f_{k-1}(\cdots (\alpha_1 f_{1}(\mathbf{X})+(1-\alpha_1)g_1(\tilde{\mathbf{A}}\mathbf{X})))\\
&\mathbf{W}_k+(1-\alpha_{k-1})\tilde{\mathbf{A}}g_{k-1}(\tilde{\mathbf{A}}\mathbf{X})\mathbf{W}_k\\
=&\sum_{d=1}^{k-1}\left(1-\alpha_d\right)\left(\prod_{j=d+1}^{k-1}\alpha_j\right)\tilde{\mathbf{A}}^{k+1-d}\mathbf{X}\mathbf{W}_d^g\\
&\left(\prod_{i=d+1}^{k}\mathbf{W}_i\right)+\left(\prod_{d=1}^{k-1}\alpha_d\right)\tilde{\mathbf{A}}^{k}\mathbf{X}\left(\prod_{i=1}^{k}\mathbf{W}_i\right),
\end{array}
$$
where $\tilde{\mathbf{A}}\mathbf{X}\mathbf{W}_k^g=g_{k}(\tilde{\mathbf{A}}\mathbf{X})$ and $\tilde{\mathbf{A}}$ denotes a filter.

Thus, Theorem~\ref{Thm1} is readily proved.
\end{proof}

As a consequence, our deep extension essentially implies multiple (variational) GAEs with different depths. In practical implementation, the combination coefficients in our models are all trainable, making our extensions easy to train.

\section{Conclusion}\label{Conclusion}
This paper aims to overcome the limitation of current SOTA methods that can only use shallow GAEs effectively. By seamlessly integrating the graph structure and node features and deeply fusing standard AEs and graph AEs, we develop an effective deep extension for shallow GAEs on link prediction. Our innovative extension leverages the joint optimization of standard and graph AEs; it also exploits a residual connection in graph AEs from the low-dimensional compact representation to the co-embedding of adjacency information and node features in graph-structured data. Further, we take advantage of a regularization technique of using the identity matrix in our proposed models. Extensive experiments on various datasets empirically demonstrate the competitive performance of our deepened graph models on link prediction. In addition, although our current extension is for the encoder, whether it is applicable to deepening the decoder will be further investigated in our future work.  

\section*{Ethical Statement}
This paper focuses on deepened graph auto-encoders for link predication. There are no ethical issues and negative societal impacts of the proposed technique.

\section*{Acknowledgments}
This work was partially supported by the NIH grants R21AG070909, R56NS117587, R01HD101508, and ARO W911NF-17-1-0040. 

\bibliographystyle{named}
\bibliography{ijcai22}

\end{document}